\documentclass[runningheads]{llncs}
\usepackage[T1]{fontenc}
\usepackage{graphicx}
\usepackage{graphicx}
\usepackage{tabularx}
\usepackage{hyperref}
\usepackage{color,soul}
\usepackage{multirow}
\usepackage{amssymb}
\usepackage[misc]{ifsym}

\begin{document}
\title{SepFormer: Coarse-to-fine Separator Regression Network for Table Structure Recognition}
\titlerunning{SepFormer: Coarse-to-fine Separator Regression Network}
%

\author{Nam Quan Nguyen\inst{1} \and
Xuan Phong Pham\inst{1} \and
Tuan-Anh Tran\inst{1,2} (\Letter)}
\authorrunning{N.Q. Nguyen et al.}
%
\institute{Viettel Artificial Intelligence and Data Services Center, Viettel Group, Lot D26 Cau Giay New Urban Area, Cau Giay District, Hanoi, Vietnam \email{ngnamquan@gmail.com, phongpx.1603@gmail.com} \and
Faculty of Computer Science $\&$ Engineering, Ho Chi Minh City University of Technology (HCMUT), VNU-HCM, Ho Chi Minh City, Vietnam \email{trtanh@hcmut.edu.vn}
}
\maketitle              
\begin{abstract}
The automated reconstruction of the logical arrangement of tables from image data, termed Table Structure Recognition (TSR), is fundamental for semantic data extraction. Recently, researchers have explored a wide range of techniques to tackle this problem, demonstrating significant progress. Each table is a set of vertical and horizontal separators. Following this realization, we present SepFormer, which integrates the split-and-merge paradigm into a single step through separator regression with a DETR-style architecture, improving speed and robustness. SepFormer is a coarse-to-fine approach that predicts table separators from single-line to line-strip separators with a stack of two transformer decoders. In the coarse-grained stage, the model learns to gradually refine single-line segments through decoder layers with additional angle loss. At the end of the fine-grained stage, the model predicts line-strip separators by refining sampled points from each single-line segment. Our SepFormer can run on average at $25.6$ FPS while achieving comparable performance with state-of-the-art methods on several benchmark datasets, including SciTSR, PubTabNet, WTW, and iFLYTAB.


\keywords{Table structure recognition  \and Line regression \and Detection transformer.}
\end{abstract}
\section{Introduction}
The automation of document processing workflows is increasingly displacing manual data entry, a process inherently susceptible to temporal inefficiencies and human error, thereby accelerating digital transformation solutions. Within this domain, tabular structures serve as a popular mechanism for the concise and organized presentation of data. Table Structure Recognition (TSR) aims to accurately reconstruct the logical organization of tables embedded within document images, translating the visual layout into machine-interpretable representations, typically expressed as logical coordinate systems or structured markup languages. However, the complexity of table layouts, which encompasses diverse variations in size, various visual styles, and implicit structural components, presents a significant challenge to the robust and precise reconstruction of tabular structures.

Early foundational TSR research, such as \cite{wang2004table,chen2012model,ref_mixture}, focused on extracting and analyzing rules-based features based on visual patterns (for example, spaces, connected components, text blocks, and vertical/horizontal lines). These methods usually require deep insight into the dataset to design heuristic logic and manually tune parameters. Although these methods require significant effort, they suffer from reliance on extensive assumptions about the quality of the image and table structure, leading to poor generalization. Over the last few years, many deep learning methodologies have achieved notable improvements in both accuracy and scalability compared to traditional methods. These approaches can be divided into three categories based on modeling the TSR problem: bottom-up methods, methods based on markup language, and methods based on table component extraction. For example, bottom-up methods \cite{rethinking_graph,show_read_reason,ncgm}  leverage supplementary textual information and formulate table structure recognition as a graph prediction task. In markup language-based approaches \cite{zhong2020image,kawakatsu2024multi,deng2019challenges}, the table is treated as sequential data (e.g., HTML or LaTeX) and employs autoregressive models to directly predict the table structure from the input image. The table component-based method \cite{pubtable1m,tsrformer,semv3,gridformer} focuses on predicting different elements of the table structure, such as separators, cells, rows, and columns.

In this paper, we propose a single-shot separator detection-based model to balance processing time and performance. Our method, named \textbf{Sep}arator regressing with Trans\textbf{Former} (SepFormer), directly predicts separators by a regression approach from table images. In this way, our approach can eliminate the need for segmentation masks in the split stage and the ROIAlign function in the merge module. Usually, segmentation tasks will involve low-level features at high resolution and need additional post-processing steps to extract lines from the masks. In search of a real-time solution, removing ROIAlign significantly reduces computational resources. SepFormer comprises two key components: 1) CNN backbone and hybrid encoder, adapted from RT-DETR \cite{rtdetr}, to extract and enhance features from the original image with high efficiency and speed; 2) Dual two-stage decoder branches with supervision to predict coarse-to-fine row and column separators. Specifically, the first decoder predicts single-line separators, in format $4D$ coordinates $\{(x_i,y_i)|i=1,2\}$. Subsequently, the $P$ points are evenly sampled between $(x_1, y_1)$ and $(x_2, y_2)$, serving as reference points for the second decoder to regress line-strip separators, in the format $\{(x_i,y_i)|i=1, \ldots, P\}$.

Extensive experiments are conducted to verify the effectiveness of our proposed method. With a simple pipeline, our method has achieved state-of-the-art or comparable performance on public benchmarks, including SciTSR\cite{scitsr}, PubTabNet \cite{zhong2020image}, WTW \cite{wtw} and iFLYTAB \cite{semv2}. The main contributions of this paper are summarized as follows:
\begin{itemize}
    \item A coarse-to-fine approach for TSR to regress row/column separators and reconstruct the table in a single shot. We take advantage of real-time architecture techniques to achieve superior processing time.
    \item A comparable performance on multiple complicated table datasets with current SOTAs. We conducted experiments on almost all challenging TSR datasets to demonstrate the robustness of SepFormer.
    
\end{itemize}

%
%

\section{Related Works}
\subsection{Table Structure Recognition}
\subsubsection{Table component extraction based methods.}
These methods involve a two-stage process: first, the fundamental elements of the table are identified, and then connected to reconstruct the table. One strategy, as seen in \cite{tabstruct}, uses object detection to pinpoint individual cells and subsequently creates a graph to obtain the structure of the table. The most challenging for cell detection are empty cells, which can confuse the detection process due to their lack of visual features. Alternatives, such as those in \cite{deepdesrt,pubtable1m,tablenet}, focus on recognizing rows and columns, deriving cell locations from their intersections. More recently, a shift towards identifying row and column separators has proven more effective than direct row/column region detection. This separator-based method, often employing a "split-and-merge" framework, initially establishes the core grid of cells and then recovers merged cells to complete the table structure \cite{splerge,rethinking,sem,semv3,trust,tsrformer,robust_dq,grab,gridformer,trace,rtsr}. Separator-based recognition has demonstrated significant robustness in table structure recognition; therefore, optimizing inference speed is crucial for practical applications.
\subsubsection{Markup language-based methods.}
Methods in this type treat TSR as an image-to-text generation problem to convert visual table representations into textual descriptions that encapsulate both the table layout and its cell contents. These textual outputs can be encoded using markup languages like HTML \cite{li2020tablebank,zhong2020image,kawakatsu2024multi} or mathematical notation such as LaTeX \cite{deng2019challenges}, with both formats mutually convertible. A notable drawback of these techniques is their computational demand, requiring substantial time and memory, particularly when processing tables with numerous cells. To address this concern about efficiency, \cite{lysak2023optimized} introduced a novel markup language, OTSL, aimed at enhancing performance by optimizing the HTML representation of table structures, resulting in improved accuracy and faster inference times. However, even with these advances in processing speed, the general methodology, including this specific optimization, still falls short of meeting the real-time requirements of practical applications.
\subsubsection{Bottom-up methods.}
Another set of techniques models the table structure using graphs, where textual elements, such as words or cell contents, serve as the graph's nodes. Graph neural networks are then used to determine the connections between selected node pairs, identifying whether they belong to the same cell, row, or column. Some earlier methods, including those described in \cite{scitsr,rethinking_graph,show_read_reason,ncgm}, rely on supplementary data on text location and content. This information might be derived from PDF metadata or obtained through optical character recognition (OCR) software. More recent advances, such as \cite{res2tim,e2egraph}, integrate a text detection component directly into the system, merging it with graph convolutional networks (GCNs) to form a unified processing flow.

\subsection{TSR with deformation image}
Deformed images present significant challenges for TSR. These deformations can manifest as rotations, skews, warps, and perspective distortions. Several methodologies have been proposed to model and overcome these deformations. GrabTab \cite{grab} employs cubic Bézier curves to represent curved separators, effectively modeling individual separator geometries. GridFormer \cite{gridformer} utilizes a vertex-edge prediction framework to address distorted table structures. Segmentation mask-based approaches, such as SEMv2 \cite{semv2}, TRACE \cite{trace}, and RTSR \cite{rtsr}, extract separator polygons from predicted masks. While TSRFormer \cite{tsrformer}, SEMv3 \cite{semv3}, and TSRFormer with DQ-DETR \cite{robust_dq} employ regression techniques, dividing each separator into $P$ equidistant points.

\section{Method}

\begin{figure}[!t]
\includegraphics[width=\textwidth]{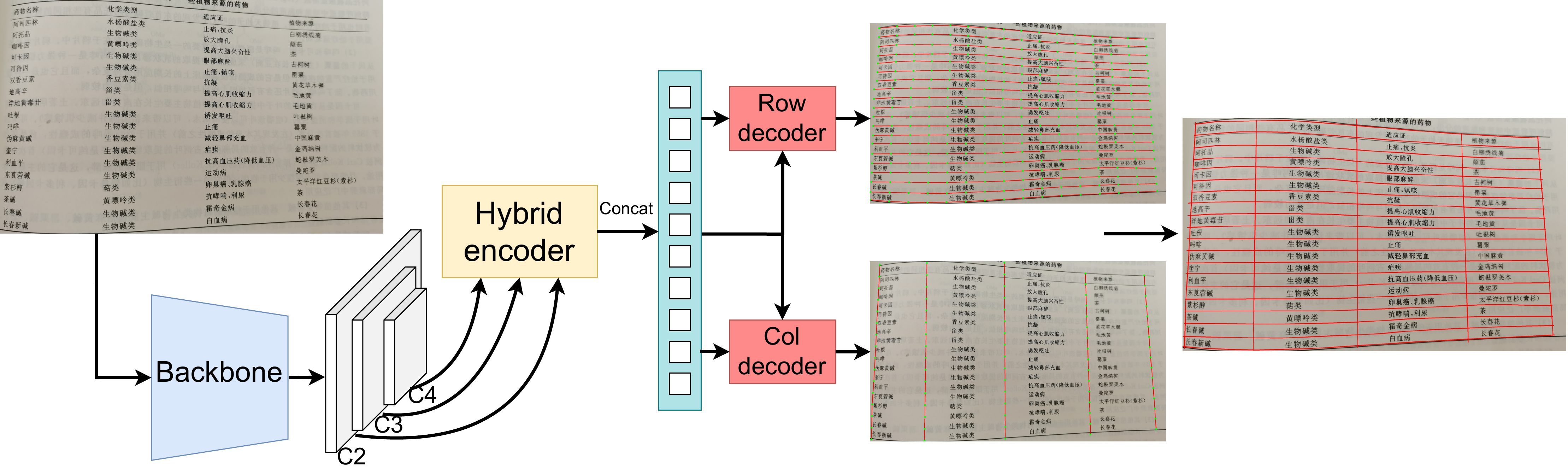}
\caption{Overview of the proposed SepFormer.} 
\label{system}
\end{figure}
\subsection{Overview}
Fig. \ref{system} illustrates the model architecture. SepFormer comprises a CNN backbone that extracts features from the input image $X\in R^{H\times W\times 3}$, a hybrid encoder to refine the CNN features, and dual decoders corresponding to the separator heads of rows and columns. Specifically, we use Resnet-34 \cite{resnet} to generate multiscale feature maps and select the last three levels of features, ${C2, C3, C4}$, corresponding to steps $8$, $16$, and $32$ relative to image resolution. Instead of employing a Feature Pyramid Network (FPN) to fuse multiscale features, we adopt an efficient method from RT-DETR \cite{rtdetr}. This approach feeds only the highest-level feature map, $C4$, into a transformer encoder for intrascale feature interaction. Then, a cross-scale feature fusion module will be used to enrich adjacent scale features. Subsequently, a cross-scale feature fusion module enriches adjacent scale features. After enhancing the information through the hybrid encoder, the three feature maps $\{M_1, M_2, M_3\}$ are flattened and concatenated into a single sequence $M\in R^{S\times C}, M=\{M_i| i=1,2,3\}$, while $S=\sum^{3}_{i=1}
 \frac{H\times W}{2^{6i}}$ and $C$ is the number of channels, set to 256 in our experiments. The sequence $M$ is then fed into dual decoders to output column and row separators. To explain these two modules in detail, we will use the row separation prediction branch as a reference. The details of the decoder will be discussed in the subsequent section. 
 
\subsection{Coarse-to-fine decoder}
\begin{figure} [!t]
\includegraphics[width=\textwidth]{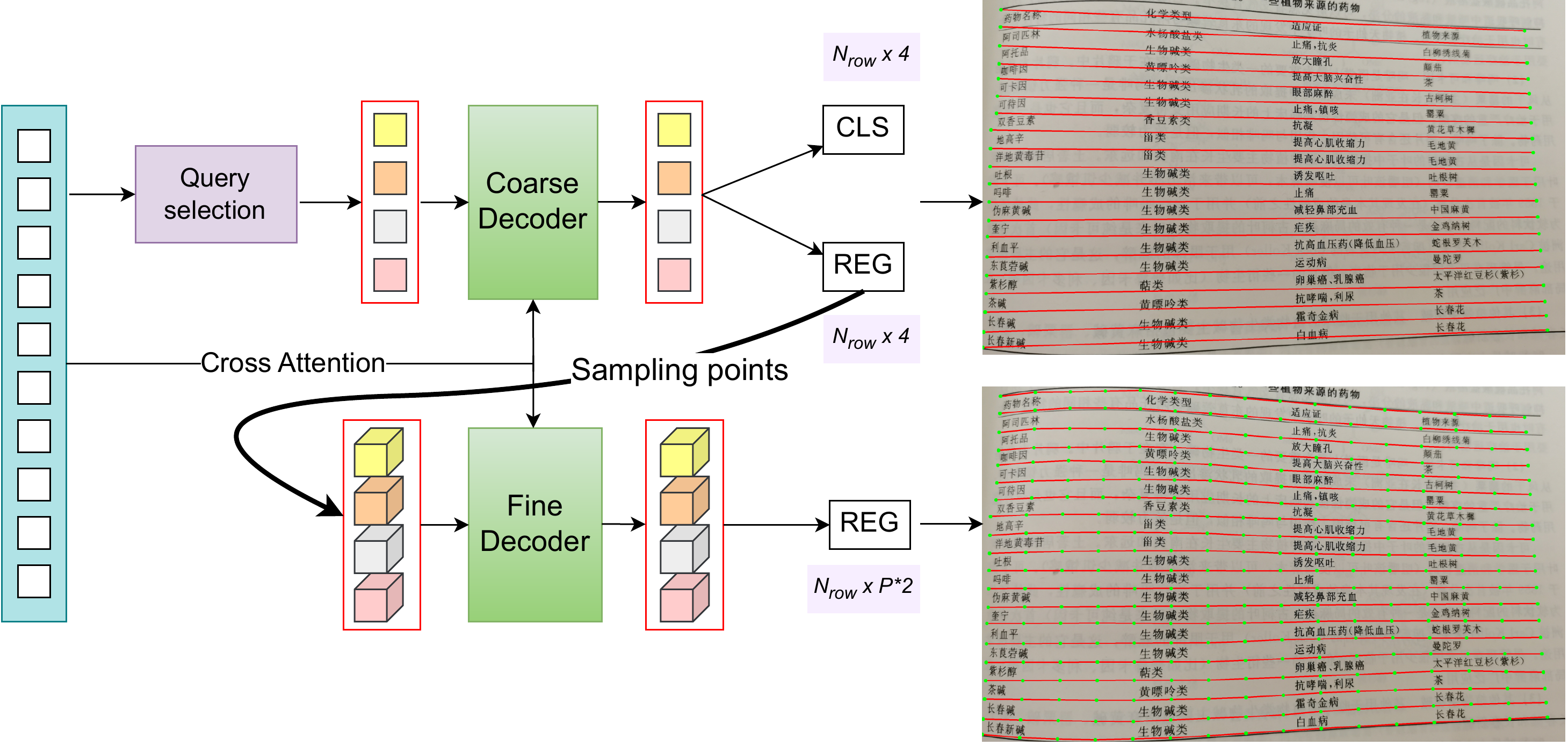}
\caption{The architecture of the row decoder for row separator prediction.} \label{dual_decoder}
\end{figure}
\begin{figure} [h]
\includegraphics[width=\textwidth]{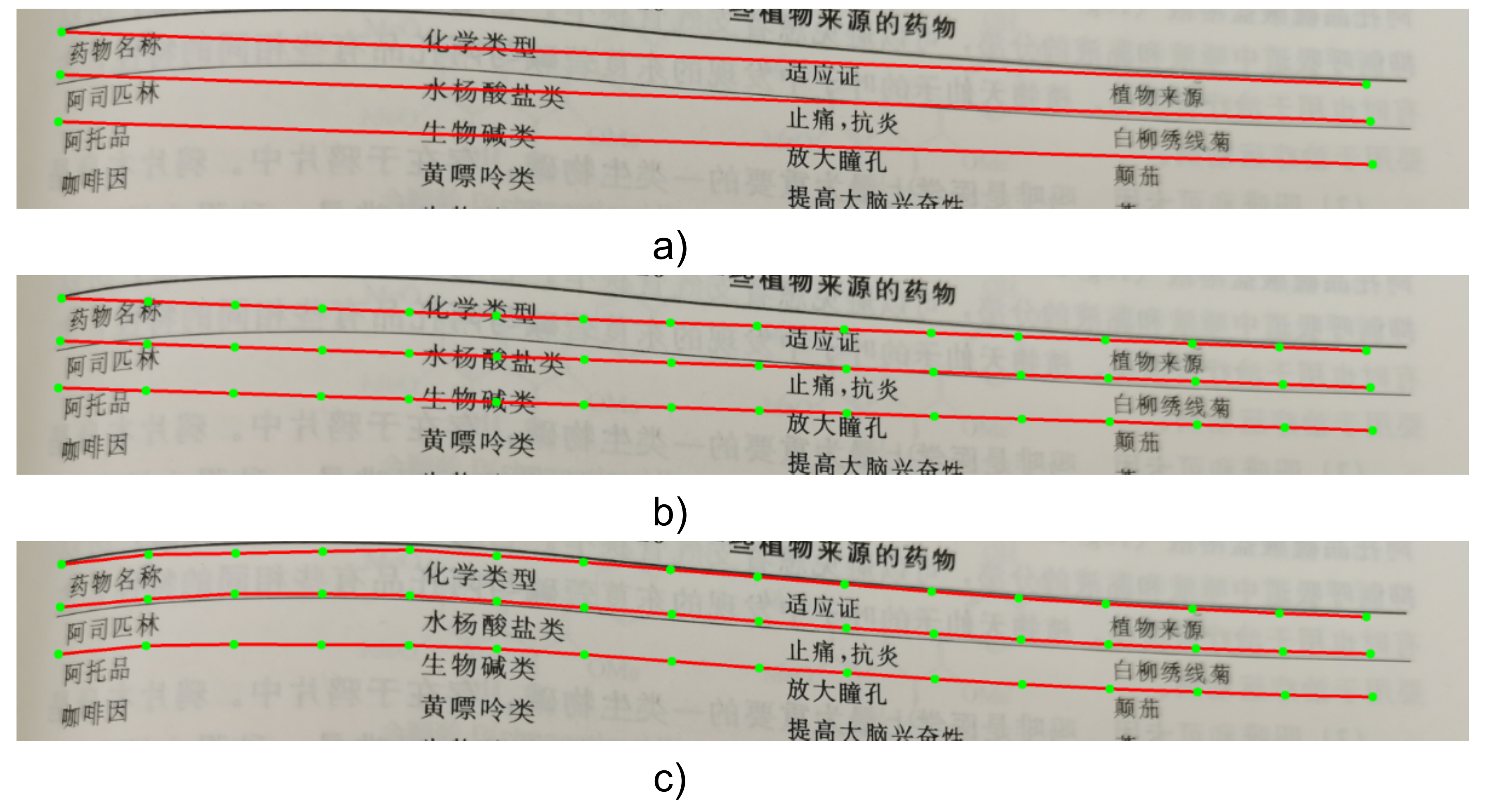}
\caption{Coarse-to-fine results in each step. a) is the single-line result from the coarse phase with line prediction; b) contains $P$ points evenly sampling from a line in a); c) is the line-strip result after refining with the fine phase.} \label{sampling}
\end{figure}

Fig. \ref{dual_decoder} shows the proposed two-stage decoder architecture. Initially, a coarse decoder performs single-line regression. Subsequently, evenly spaced points are sampled, and a fine decoder refines the prediction through line-strip regression. The sequential results of this process are visualized in Fig. \ref{sampling}. 
\subsubsection{Query selection.} Following the two-stage deformable DETR scheme \cite{deformabledetr}, we employ a query selection module to provide high-quality queries to the decoder. A learning process is applied to each element of the sequence $M$ using a detection head. The ground truth consists of the coordinates of single-line separators.  Unlike \cite{deformabledetr}, which generates $4D$ bounding box proposals $\{x_i, y_i, 2^{i-1} \times s, 2^{i-1}\times s\}$ at each pixel of feature level $i$ and object scale $s$, we adapt the approach to generate line proposals. We believe that straight-line anchors will enhance the model's awareness of line prediction. Therefore, the row proposals are defined as $\{x_i, y_i, x_i + 2^{i-1}\times s, y_i\}$, and the column proposals as $\{x_i, y_i, x_i, y_i + 2^{i-1}\times s\}$.  The detection head comprises a 3-layer feedforward network (FFN) for separator regression and a linear projection for separator binary classification. The top $K_{row}$ separator coordinates with the highest classification score $c_k$ are used to initialize reference points in the decoder.  
\subsubsection{Coarse Decoder.}
The queries in the decoder consist of two parts: content queries (decoder embeddings) and positional queries (generated from reference points). We follow \cite{dino} to get the embedding in sequence $M$ at the position of the top-$K_{row}$ as content queries. We employ three deformable transformer decoder layers \cite{deformabledetr} with iterative layer-by-layer refinement to obtain output embeddings. Separators are obtained after this is $\{(c^k, l^k)|k=1,...,K_{row}\}$, where $l^k=\{(x^k_i,y^k_i)|i=1,2\}$ is a single-line coordinate and $c^k$ is the classification probability  corresponding to the $k-th$ separator. In Section \ref{matching}, we describe the separator matching strategy that matches the fixed number predictions $K_{row}$ with the arbitrarily numbered ground truth to make the network trainable from end to end.
\subsubsection{Sampling points.}
This function only applies to a single-line separator in optimal matching in training and is filtered in the inference phase. For each chosen separator, we sample evenly $P=16$ points in each line $\{(x_1,y_1),(x_2,y_2)\}$ with following formulate,
\begin{equation}
    Sampling=\{(1-\frac{t}{P}) * (x_1,y_1) + \frac{t}{P} * (x_2,y_2)|t=1,...,P\}
\end{equation}
\subsubsection{Fine decoder.} 
We keep query embedding and memory similar to the previous decoder, only reference points are used after sampling. We also use three iteratively refined deformable transformer decoders. The output embedding is passed to a 3-layer FFN to regress the coordinates of the line trip separators. In this stage, there is only one regression head to refine the line strip separators. The coordinate of $k-th$ line-strip separator $ls^k=\{(x^k_i,y^k_i)|i=1,...,P\}$ is appended to the separator information.

\subsection{Separator matching}
\label{matching}
\subsubsection{Bipartite Matching.} Let denote $\{(c_{gt}^k,l_{gt}^k,l_{gt}^k)|k=1,...,N_{row}\}$ is set of $N_{row}$ label row separators. We optimize a bipartite matching objective in a permutation function $\sigma$ that maps prediction indices $K_{row}$ to potential target indices $\{1, \ldots, N_{row}\}$

\begin{equation}
\mathcal{L}_{match} = \sum^{N_{row}}_{i=1}(\lambda_{coord}||l_{gt}^i - l^{\sigma(i)}|| + \lambda_{cls}c^{\sigma(i)})
\end{equation}
\begin{equation}
\hat{\sigma} = argmin_{\sigma}\mathcal{L}_{match}
\end{equation}

$\mathcal{L}_{match}$ is a function that measures the matching cost, taking into account both distance and confidence with the balancing coefficients $\lambda_{coord}, \lambda_{cls}$. The optimal permutation $\hat{\sigma}$ is calculated using the Hungarian algorithm. During the inference stage, we filter the predictions of the single-line separator by setting a fixed threshold $\tau_{row}$ based on the confidence.

\subsubsection{Loss function}
After finding the optimal permutation, we can calculate the loss function, which consists of four parts: a separation label classification loss, an angle loss, and two coordinate regression losses.  The row label classification loss is a standard binary cross-entropy:

\begin{equation}
\mathcal{L}^{row}_{cls} = -\frac{1}{N_{row}}\sum^{N_{row}}_{n=1}{c^n_{gt}.log({c}^n)+(1-c^n_{gt}).log(1-{c}^n)}
\end{equation}

We use the cosine similarity function $cos(\cdot)$ to estimate the difference between the prediction and the single-line label. The term $||c^n_{gt}||\times 4$ is the penalty for short separators; this will be an illustration in Fig. \ref{angle_loss}:

\begin{equation} \label{angle_formulate}
\mathcal{L}^{row}_{angle} = \frac{1}{N_{row}}\sum^{N_{row}}_{n=1}{(1 -\frac{cos(c^n_{gt},{c}_n)}{||c^n_{gt}|| \times 4})}
\end{equation}

Two coordinates regression losses are distances based on $L_1$ for a single-line and line-strip separators:

\begin{equation}
\mathcal{L}^{row}_{line} = \frac{1}{N_{row}}\sum^{N_{row}}_{n=1}{||l^n_{gt}-l^n||}
\end{equation}

\begin{equation}
\mathcal{L}^{row}_{linestrip} = \frac{1}{N_{row}}\sum^{N_{row}}_{n=1}{||ls^n_{gt}-ls^n||}
\end{equation}

The objective function for row regression is a combination of the above losses with balance factors $\lambda_1, \lambda_2,\lambda_3, \lambda_4$:
\begin{equation}
\mathcal{L}^{row} = \lambda_1\mathcal{L}^{row}_{cls} + \lambda_2\mathcal{L}^{row}_{angle} + \lambda_3\mathcal{L}^{row}_{line} + \lambda_4\mathcal{L}^{row}_{linestrip}
\end{equation}

Complete loss of SepFormer includes row and column separator regression:
\begin{equation}
\mathcal{L}^{sep} = \mathcal{L}^{row} + \mathcal{L}^{col} 
\end{equation}

\section{Experiments}

\subsection{Datasets and metrics}
\subsubsection{SciTSR} \cite{scitsr} is a large-scale table structure recognition dataset from scientific papers. It contains $15000$ PDF tables divided into $12000/3000$ for training and testing. The author also created a SciTSR-COMP sub-dataset with $2885$ and $716$ extremely complicated tables in the training and test sets to increase the challenges. In this research, we used the SciTSR-COMP set as the primary evaluation scheme. As presented in \cite{scitsr}, the metric for this dataset is the adjacent relationship score of the cell.
\subsubsection{Pubtabnet} \cite{zhong2020image} is a large table structure recognition dataset that extracts research papers from the medical domain. This data set contains $500777/ 9115/ 9138$ documents for training/validation and testing. Because the labeling of the testing set has not yet been released, the validation set is used for the evaluation profile. This article also proposes a measure called TED to evaluate performance. However, the OCR metric is not fair when considering table restructuring. Therefore, several modified versions, such as TEDS-Struct, have been proposed and are widely used in TSR competitions. We also use this modified metric to evaluate our approach on this dataset.

\subsubsection{WTW} \cite{wtw} is a large-scale table structure recognition dataset designed to evaluate the performance of models in real-world scenarios. It encompasses 14581 wired table images collected from various sources, including photographs, scanned documents, and Web pages. The dataset is characterized by its inclusion of tables with varying styles and complexities, featuring challenging cases such as inclined, curved, occluded, and extreme aspect ratio tables. The WTW dataset is divided into $10970$ training samples and $3611$ testing samples. The metric on WTW is the cell adjacency relationship. 
\subsubsection{iFLYTAB} \cite{semv2} is a challenging TSR dataset introduced to comprehensively evaluate TSR models. It comprises a total of 17291 samples, divided into 12104 samples for training and $5187$ samples for testing. This dataset encompasses a variety of table styles collected from various scenarios, including wired and wireless tables from digital documents, as well as camera-captured photos and scanned documents.  It was created to address the complexities and diversity of table structures, making it a more demanding benchmark for TSR methods. Following \cite{semv2,semv3}, the cell adjacency relationship score was also used as an evaluation metric.

\subsection{Implementation details}
\begin{table}[!b]
\caption{Speed performance of the real-time methods on four datasets. The unit is FPS (frame per second)}. \label{tab_fps}
\centering
\begin{tabular}{m{2.75cm} m{1.75cm} m{1.75cm} m{1.75cm} m{1.75cm} m{1.5cm}}
\hline
 \hfil\bfseries Method/Dataset & \hfil\bfseries SciTSR &  \hfil\bfseries Pubtabnet & \hfil\bfseries WTW & \hfil\bfseries iFLYTAB & \hfil \bfseries Average \\ 
\hline
\hfil RTSR \cite{rtsr} & \hfil \textcolor{red}{45.8} & \hfil \textcolor{red}{43.2} & \hfil \textcolor{red}{25.5} & \hfil \textcolor{red}{22.9} & \hfil \textcolor{red}{34.4} \\
\hfil SepFormer & \hfil 28.5 & \hfil 26.9 & \hfil 25.2 & \hfil 21.7 & \hfil 25.6 \\
\hline
\end{tabular}
\end{table}
All experiments are conducted using PyTorch $2.1.2$. The training phase is on a single NVIDIA V100 32GB, while inference was executed on an RTX 3060. The weights of the ResNet-34 are initialized from the pre-trained model of ImageNet. The transformer consists of $1$ encoder, $3$ coarse decoders, and $3$ fine decoder layers with $256$ channels. We use $8$ heads and $4$ sampling points for the deformable attention module. We select the top $K_{row}=300$, $K_{col}=300$ encoder features to initialize object queries of the coarse decoder. We set an initial learning rate of $3e-5$ and schedule cosine annealing in $100$ epochs, except for Pubtabnet with $20$ epochs. In the training phase, we rescale the longer side of the table image to a number in \{$864, 896, 928, 960$\} while keeping the aspect ratio, except for all datasets. In the testing phase, we resize the longer side of each image to $896$. The matching coefficients $\lambda_{cls},\lambda_{coord}$ are set to $2, 3$; loss coefficients $\lambda_1, \lambda_2, \lambda_3, \lambda_4$ are set to $1, 1, 3, 1$. The fixed score threshold $\tau_{row}$ and $\tau_{col}$ is set to $0.95$ and $0.95$, respectively, for all experiments.

\subsection{Experiment results}
\begin{figure} [!t]
\includegraphics[width=\textwidth]{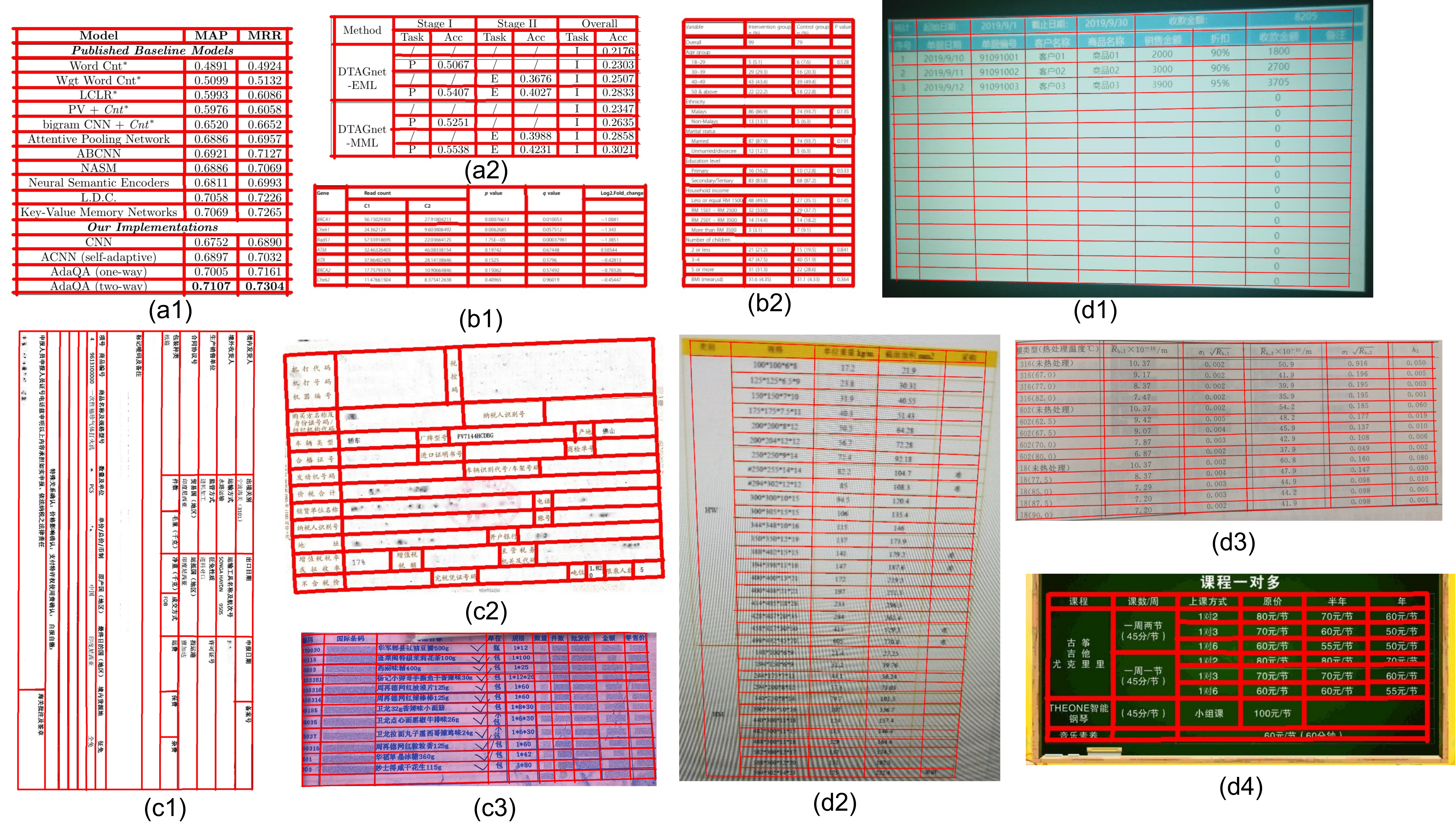}
\caption{Experiment results of SepFormer on various datasets. (a1-2) SciTSR, (b1-2) Pubtabnet, (c1-3) WTW, (d1-4) iFLYTAB.} 
\label{good}
\end{figure}

\begin{table}[!b]
\caption{Evaluation profile for the SOTA methods on the SciTSR-COMP and Pubtabnet dataset.} \label{tab_scitsr}
\centering
\begin{tabular}{c c ccc c ccc c}
\hline

\multirow{2}{*}{\bfseries Method/Dataset} && \multicolumn{3}{c}{\bfseries SciTSR-COMP} &&
{\bfseries Pubtabnet}\\
\cline{3-5} \cline{7-7} 
 && \bfseries Prec. & \bfseries Rec. & \bfseries F1. && \bfseries TEDS-S
 \\
\hline
 TabStruct-Net \cite{tabstruct} && 90.9 & 88.2 & 89.5 && - \\



  FLAG-Net \cite{show_read_reason} && 98.4 & 98.6 & 98.5 && - \\

  TSRFormer w/ DQ-DETR \cite{robust_dq} && {99.1} & {98.6} & {98.9} && \textcolor{red} {97.5} \\
  
  SEMv2 \cite{semv2} && 98.7 & 98.6 & {98.7} && \textcolor{red}{97.5} \\

  SEMv3 \cite{semv3} && 99.1 & 98.9 & {99.0} && \textcolor{red}{97.5} \\

  TRUST \cite{trust} && - & - & - && 97.1 \\
    
  GridFormer \cite{gridformer} && - & - & - && 97.0 \\

  GrabTab \cite{grab} && 98.9 & 99.4 & 99.1 && - \\

  LORE++ \cite{lorepp} && \textcolor{red}{99.4} & \textcolor{red}{99.2} & \textcolor{red} {99.3} && - \\
  
\hline
\hline
\multicolumn{1}{l}{\itshape \bfseries Real-time system} \\ 
   RTSR \cite{rtsr} &&  {98.7} & 98.0 & 98.3 && 95.7\\
   Our SepFormer &&  99.0 & 98.2 & 98.6 && 96.8 \\
\hline
\end{tabular}

\end{table}
Beyond the evaluation of four public datasets, SciTSR-COMP, PubTabNet, WTW, and iFLYTAB, our research has been directly applied to customer data within a commercial system. The visual results, shown in Fig. \ref{good}, and the performance metrics in Table \ref{tab_fps}, demonstrate the stability of the system in both computational performance and precision. Our TSR system was evaluated against current SOTA approaches using two primary criteria: time-consuming and accuracy.

Regarding processing time, Table \ref{tab_fps} confirms that our average processing time across the four datasets achieves real-time performance. Although slightly slower than RTSR \cite{rtsr}, our method offers a higher and more stable accuracy.

Concerning accuracy, we benchmarked SepFormer against other SOTA TSR techniques. These data sets are structured to progressively increase the structure of the table and the complexity of the deformation image. 
As detailed in Table \ref{tab_scitsr}, our approach exhibited competitive performance in our initial benchmark, which used datasets of scanned documents and high-resolution images. Specifically, on the SciTSR-COMP dataset, we achieved an F1-score of $98.6\%$, a marginal difference of $0.7\%$ compared to LORE++, the top performer. In the larger PubTabNet dataset, SepFormer also demonstrated an acceptable performance gap of $0.9\%$.

When tested on complex datasets featuring photo-captured images with skew, warp, and low saturation conditions, the performance of our method remained within $1.2\%$ of the top performers in WTW (Table \ref{tab_wtw}) and $0.6\%$ on iFLYTAB (Table \ref{tab_iflytab}). In particular, iFLYTAB, a newly released dataset in 2024, presents significant challenges and currently only has the results of the author group with many different versions. Even with these increased challenges, SepFormer maintains acceptable performance gaps compared to current SOTA methods.

\begin{table}[!t]
\caption{Evaluation profile for the SOTA methods on the WTW dataset.} \label{tab_wtw}
\centering
\begin{tabular}{m{5cm} m{2cm} m{2cm} m{2cm}}
\hline
 \hfil\bfseries Method &  \hfil\bfseries Prec. & \hfil\bfseries Rec. & \hfil \bfseries F1. \\ 
\hline
    \hfil Cycle-CenterNet \cite{wtw}& \hfil 93.3 & \hfil 91.5 & \hfil 92.4 \\

    \hfil TSRFormer w/ DQ-DETR \cite{robust_dq} & \hfil 94.5 & \hfil 94.0 & \hfil 94.3\\
    
    \hfil SEMv2 \cite{semv2} & \hfil {93.8} & \hfil 93.4 & \hfil {93.6}\\

    \hfil SEMv3 \cite{semv3} & \hfil {94.8} & \hfil 95.4 & \hfil \textcolor{red}{95.1} \\
        
    \hfil GridFormer \cite{gridformer} & \hfil 94.1 & \hfil 94.2 & \hfil 94.1 \\

    \hfil GrabTab \cite{grab} & \hfil \textcolor{red}{95.3} & \hfil 95.0 & \hfil \textcolor{red}{95.1} \\

    \hfil LORE++ \cite{lorepp} & \hfil 94.5 & \hfil \textcolor{red}{95.9} & \hfil \textcolor{red}{95.1} \\
    
\hline
\hline
\textit{\textbf{Real-time system}} & & & \\
\hfil RTSR \cite{rtsr} & \hfil 92.2 & \hfil 93.6 & \hfil 92.9 \\
\hfil Our SepFormer & \hfil 93.7 & \hfil {94.2} & \hfil 93.9 \\
\hline
\end{tabular}
\end{table} 

\begin{table}[!b]
\caption{Evaluation profile for the SOTA methods on the iFLYTAB dataset.}\label{tab_iflytab}
\centering
\begin{tabular}{m{5cm} m{2cm} m{2cm} m{2cm}}
\hline
 \hfil\bfseries Method &  \hfil\bfseries Prec. & \hfil\bfseries Rec. & \hfil \bfseries F1. \\ 
\hline
    \hfil SEM \cite{sem}& \hfil 81.7 & \hfil 74.5 & \hfil 78.0 \\
    
    \hfil SEMv2 \cite{semv2} & \hfil {93.8} & \hfil 93.3 & \hfil {93.5}\\

    \hfil SEMv3 \cite{semv3} & \hfil 94.4 & \hfil \textcolor{red}{94.2} & \hfil \textcolor{red}{94.4} \\
\hline
    \textit{\textbf{Real-time system}} & & & \\
    \hfil RTSR \cite{rtsr} & \hfil {91.8} & \hfil 90.5 & \hfil 91.1 \\
    \hfil Our SepFormer & \hfil \textcolor{red}{94.6} & \hfil 93.2 & \hfil 93.8 \\
\hline

\end{tabular}
\end{table}

\begin{figure} [!t]
\includegraphics[width=\textwidth]{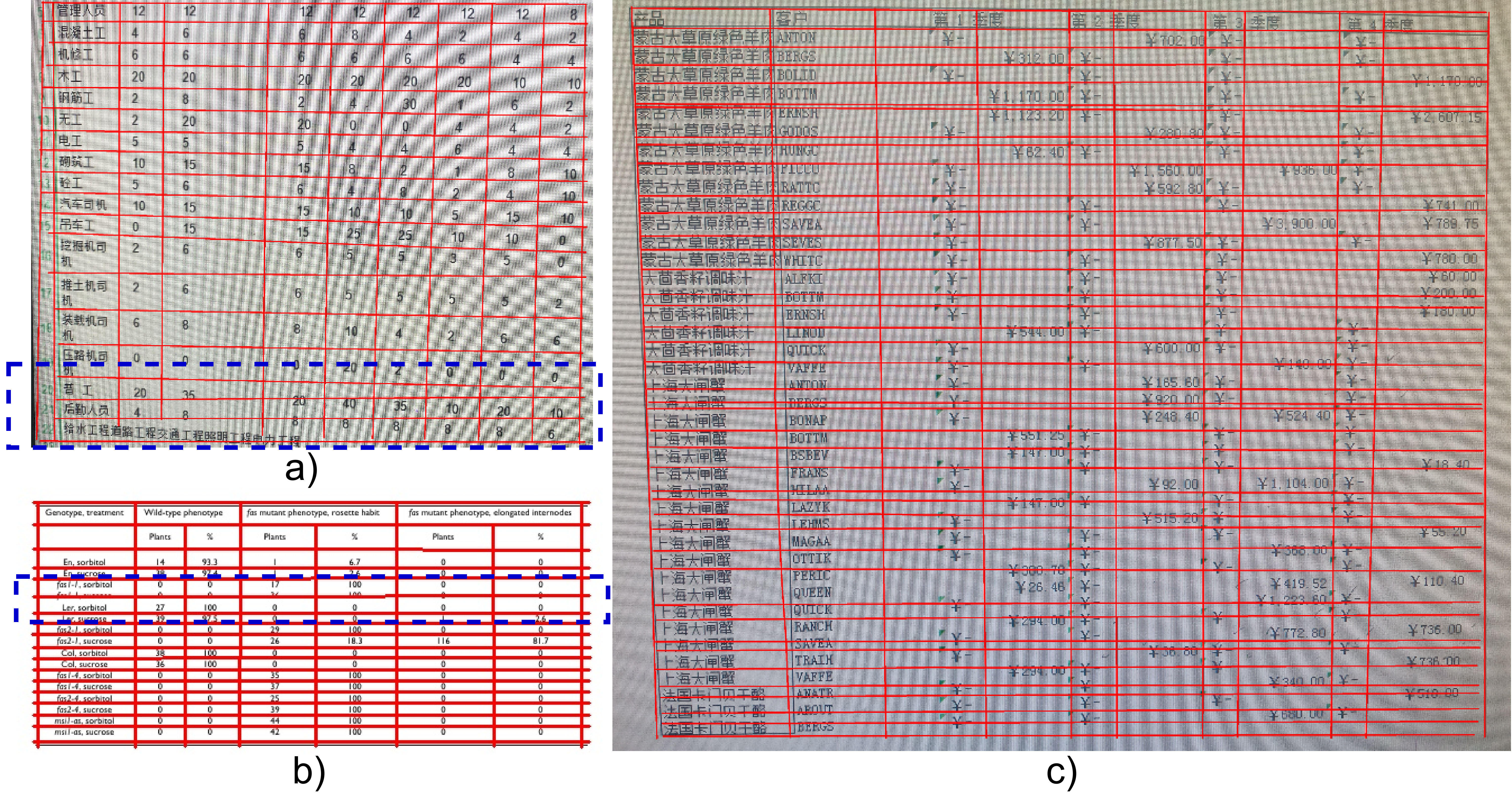}
\caption{Some failure cases from SepFormer. The blue dashed box is the error of prediction. \textbf{(Zoom in for better view)}}
\label{bad}
\end{figure}

\subsection{Result analysis}
This section analyzes the experimental results obtained from four distinct datasets: SciTSR, Pubtabnet, WTW, and iFLYTAB. The qualitative results, presented in Figures \ref{good} and \ref{bad}, highlight the stability of SepFormer, a critical factor for application research. The system demonstrates robust performance across various table complexities, including variations in the background (Fig. \ref{good}-c3-d1-d4), tilted or dewarped tables (Fig. \ref{good}-c2-d2), and tables with rotation (Fig. \ref{good}-c1). SepFormer also maintains precision on simpler table layouts (Fig. \ref{good}-a1-a2-b1). Tables with moderate row and column density consistently yield favorable results. However, excessively dense layouts, as illustrated in Fig. \ref{bad}c, can lead to over or under-detection of rows, a challenge even for human interpreters.
Furthermore, errors can arise with heavily warped images, as seen in Fig. \ref{bad}a, where text lines exhibit varying degrees of warping within the same image. In other failures (Fig. \ref{bad} b), the model outputs a single line instead of the expected two separations, which appears to approximate their midpoint. We hypothesize that these errors come from the matching process. Specifically, the $L_1$ distance criterion used for single-line matching may lack sufficient discrimination. Consequently, the matching process exhibits imprecision and a tendency to confusion when presented with closely spaced separators. 

Regarding processing time, many contemporary Table Structure Recognition (TSR) algorithms prioritize accuracy, resulting in performance rates of approximately 2-3 images per second (FPS). In contrast, our solution, SepFormer, is designed to optimize image processing speed for practical large-scale applications. As shown in Table \ref{tab_fps}, SepFormer consistently achieves real-time performance, with an average of 25.6 FPS in various data sets. Although RTSR, a segmentation-based approach, exhibits faster processing times than our regression-based method, it struggles to improve accuracy with increasing data volume. SepFormer, in contrast, overcomes this limitation and effectively handles slightly curved lines.

A trade-off between computation time and accuracy is evident across all methods. SepFormer achieves an accuracy comparable to the best solutions while offering significantly superior processing times, typically 2-10 times faster. For instance, on the SciTSR-COMP and WTW datasets, LORE++ often achieves slightly higher accuracy ($0.7\%$-$1.2\%$ better), but its average processing time is approximately 10 times slower than SepFormer's. SepFormer's average processing speed of 28.5 FPS substantially outperforms LORE++'s 2.5 FPS, as documented in \cite{lorepp}. TRUST \cite{trust}, a nonreal-time solution, achieves the fastest processing time among its counterparts (10 FPS), but is only about $0.2\%$ more accurate than SepFormer in the Pubtabnet dataset.

\subsection{Ablation study}

In the ablation study, we conducted multiple experiments to evaluate the impact of our proposal modules. All runs were performed on the iFLYTAB dataset.

\begin{table} [!t]
\caption{The ablation experiments of the two-stage decoder module. "SL-M" refers to single-line matching, while "LS-M" represents line-strip matching.} \label{tab_decoder}
\centering
\begin{tabular}{m{3cm} m{1.5cm} m{1.5cm} m{1.5cm} m{1.5cm} m{1.5cm}}
\hline
 \hfil\bfseries Decoder & \hfil\bfseries SL-M & \hfil\bfseries LS-M & \hfil\bfseries Prec. & \hfil\bfseries Rec. & \hfil \bfseries F1. \\ 
\hline
    \hfil one-stage 3-layer & \hfil \checkmark & & \hfil 91.5 & \hfil 90.1 & \hfil 90.8 \\

    \hfil one-stage 6-layer & & \hfil \checkmark &\hfil {90.1} & \hfil 89.3 & \hfil {89.7}\\

    \hfil one-stage 6-layer & \hfil \checkmark & \hfil \checkmark &\hfil {91.3} & \hfil 89.7 & \hfil {90.5}\\

    \hfil one-stage 6-layer & \hfil \checkmark & &\hfil {91.9} & \hfil 90.2 & \hfil {91.0}\\

\hline
    \hfil two-stage 3-layer & \hfil \checkmark & \hfil \checkmark &\hfil {92.7} & \hfil {91.1} & \hfil {91.9}\\
    
    \hfil two-stage 3-layer & \hfil \checkmark & &\hfil \textcolor{red}{94.6} & \hfil \textcolor{red}{93.2} & \hfil \textcolor{red}{93.8}\\

\hline
\end{tabular}
\end{table}

\subsubsection{Two-stage decoder.}
We investigate the effectiveness of decoupling two-level decoding in single-line and line-strip separators for each branch. Specifically, we implement a variant that directly predicts the line strip using a decoder and evaluate it on decoders with 3 and 6 layers. As shown in Table \ref{tab_decoder}, the two-stage decoder design with 3 layers in each stage achieves better performance than the unified design. The results show a performance improvement of $2.4\%$ compared to the 6-layer one-stage decoder and $3.0\%$ compared to the 3-layer decoder. This experiment suggests that optimizing each step enhances the model's ability to predict challenging tasks more effectively. In addition, we experiment to examine the optimal matching strategy. We introduce an additional criterion based on the line-strip distance, computed as the average $L_1$ distance of each ordered pair of points in two line-strip separators. In both unified and decoupled decoders, the incorporation of line-strip matching degrades the model performance. We find that using only single-line matching allows the model to select a better matching strategy through the Hungarian algorithm.

\begin{table}
\caption{The ablation experiments of angle loss.}\label{tab_angle}
\centering
\begin{tabular}{m{3cm} m{2cm} m{2cm} m{2cm}}
\hline
 \hfil\bfseries Angle loss &  \hfil\bfseries Prec. & \hfil\bfseries Rec. & \hfil \bfseries F1. \\ 
\hline
    \hfil No & \hfil 94.4 & \hfil 92.9 & \hfil 93.6 \\

\hline
    \hfil Yes & \hfil \textcolor{red}{94.6} & \hfil \textcolor{red}{93.2} & \hfil \textcolor{red}{93.8} \\
\hline

\end{tabular}
\end{table}

\subsubsection{Angle loss.} 
\begin{figure}
\includegraphics[width=\textwidth]{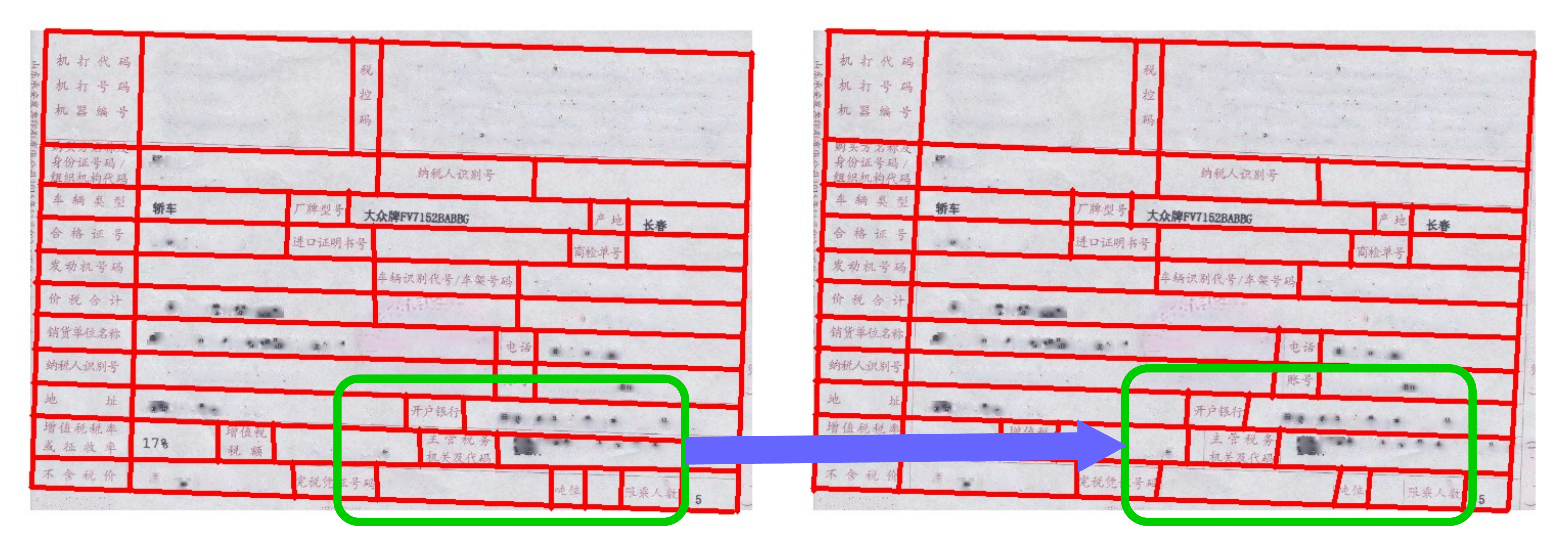}
\caption{Comparison of results with and without angle loss. The left image represents a result trained with angle loss, while the right image represents a result trained without it.}
\label{angle_loss}
\end{figure}

The impact of angle loss is presented in Table. \ref{tab_angle}. Although the improvement is marginal ($0.2\%$ F1 score), the loss of angle improves the consistency of the model in predicting short separators (Fig. \ref{angle_loss}). This is why we introduce the term $||c_{gt}^n|| \times 4$ in the denominator of the angle loss formulation (Equation \ref{angle_formulate}), increasing the penalty for deviations in the short-length separator predictions.

\section{Conclusion}
This paper introduces SepFormer, a novel real-time table structure recognition method that uses a regression approach of row and column separators. Our method employs a coarse-to-fine decoder, enabling separator recognition from single-line to line-strip levels. We conducted comprehensive experiments across diverse and challenging table datasets, including photo and scanned images, bordered and borderless tables, and flat and distorted layouts, to validate our approach. The experimental results demonstrate that SepFormer outperforms existing real-time methods while achieving performance comparable to that of state-of-the-art techniques. Although surpassing non-real-time solutions in accuracy remains a long-term objective, SepFormer demonstrates significant potential in this direction. The pursuit of faster and lighter models is a clear and essential trend, particularly when dealing with large datasets within infrastructure and resource constraints. For future work, we intend to explore line-strip regression to further enhance the stability and robustness of our performance.

\section{Acknowledgments}
We acknowledge Ho Chi Minh City University of Technology (HCMUT), VNU-HCM for supporting this study.

%
%
%
%

\end{document}